\documentclass[10pt,twocolumn,letterpaper]{article}

\usepackage{wacv}
\usepackage{times}
\usepackage{epsfig}
\usepackage{graphicx}
\usepackage{amsmath}
\usepackage{amssymb}
\usepackage{algorithm,algorithmic}
\usepackage{subfigure} 

\def\wacvPaperID{****} 

\wacvfinalcopy 

\begin{document}


\title{Few shot clustering for indoor occupancy detection with extremely low-quality images from battery free cameras}

\author{Homagni Saha\\
Iowa State University\\
Department of Mechanical Engineering\\
{\texttt\small hsaha@iastate.edu}\\
\and
Sin Yong Tan\\
Iowa State University\\
Department of Mechanical Engineering\\
{\texttt\small tsyong98@iastate.edu}\\
\and
Ali Saffari\\
University of Washington\\
Department\\
{\texttt\small saffaria@uw.edu}\\
\and
Mohamad Katanbaf\\
University of Washington\\
Department\\
{\texttt\small katanbaf@uw.edu}\\
\and
Joshua R. Smith\\
University of Washington\\
Department\\
{\texttt\small jrs@cs.washington.edu}\\
\and
Soumik Sarkar\\
Iowa State University\\
Department of Mechanical Engineering\\
{\texttt\small soumiks@iastate.edu}\\
}

\maketitle

\begin{abstract}
Reliable detection of human occupancy in indoor environments is critical for various energy efficiency, security, and safety applications. We consider this challenge of occupancy detection using extremely low-quality, privacy-preserving images from low power image sensors. We propose a combined few shot learning and clustering algorithm to address this challenge that has very low commissioning and maintenance cost. While the few shot learning concept enables us to commission our system with a few labeled examples, the clustering step serves the purpose of online adaptation to changing imaging environment over time. Apart from validating and comparing our algorithm on benchmark datasets, we also demonstrate performance of our algorithm on streaming images collected from real homes using our novel battery free camera hardware. 
\end{abstract}
\section{Introduction}
Indoor occupancy detection is fundamental to enabling a wide range of applications such as indoor home automation, activity recognition, patient health monitoring, indoor environment regulation, and most importantly energy savings in smart homes. In most residential applications, a variety of passive sensing technologies have been used to infer occupancy state of the room. In this regard, image based occupancy sensing is generally avoided to ensure privacy of its occupants and avoid computationally expensive image/video processing algorithms that might be hardware intensive~\cite{tan2019publication,tan2019granger,tang2019indoor}. However, this modality of information is invaluable in providing instantaneous occupancy status of an indoor space with high confidence. To make significant inroads into the market and have substantial impact on energy savings, current research is focusing on low cost, low power embedded devices to be used for efficient occupancy sensing. However, such devices may not provide the high-quality images. Although they may help in maintaining occupant privacy, it is a hard technical challenge to achieve high occupancy detection performance with low-quality images. 
\newline
From a computer vision perspective, generic person detection methods can be compute intensive in terms of frame rate and have deteriorated accuracy in nonstandard image collection conditions when deployed~\cite{mehmood2019object,zengeler2019evaluation}. On the other hand, simple frameworks may have generalization issues and may need to be trained on specific image settings to obtain higher accuracy. This increases commissioning and maintenance cost that could be prohibitive in a application sector such as building energy management. 
Image based approaches to detect occupancy have been proposed in several works~\cite{callemein2019anyone,acharya2018real,zou2017occupancy}. However, most of these approaches have the problem of acquiring zone specific labeled data, using expensive person detection frameworks, or generating surrogate labels by prediction from superior object detection frameworks. 
Moreover, lowering labeling cost and adaptability of the algorithm to different scenarios were barely studied. 
\newline
In this paper, we address these issues by proposing a label-efficient and adaptive human occupancy detection framework that can handle extremely low-quality images. We build up on the recent progress made on the few shot learning approaches for image classification that work with extremely small number of labels. We incorporate a novel metric learning based few shot approach in occupancy estimation that has low commissioning and maintenance cost. To cope with varying image quality over time, we design an expectation maximization (E-M) style clustering algorithm that serves the purpose of online adaptation. We validate our algorithm on a novel battery free embedded hardware that can only provide extremely low-quality images. 

\noindent \textbf{Contribution:} Our specific contributions in this paper are:
\begin{enumerate}
    \item We propose a few shot learning framework that enables relatively simpler convolutional neural network (CNN) architectures for indoor human occupancy detection using extremely low-quality images. 
    \item We present a few shot clustering technique as an online adaptation approach for the indoor occupancy detection framework. In our framework, users only need to provide a few starting images to set the semantics for clustering and then the algorithm learns to cluster instances into occupied and unoccupied classes in an unsupervised fashion from streaming videos under various situations. 
    \item We compare our technique with other competitive techniques and show better performance on benchmark datasets. 
    \item We demonstrate our algorithm on images captured by a low cost battery free camera hardware and release the images as a new publicly available dataset.
\end{enumerate}
\section{Related work}
Following are the major available options in the realm of few shot learning:
\\
\textbf{Model Agnostic Meta Learning (MAML) strategy} - Also recognized as a learning to learn approach. Several innovative techniques have been proposed as in \cite{finn2017model,ravi2016optimization}. Here the aim is to come up with a neural network weight configuration that can be easily finetuned to any new task with lower number of training iterations. We do not go with this approach as we feel it is an unnecessarily expensive approach to detect binary occupied or unoccupied instances in rooms. Meta learning requires huge number of samples (albeit different class) to learn initial weight configuration and proposed neural network models are not always lightweight because it needs to take care of generalizing to any new scenario.
\\
\textbf{Memory based learning approaches} - Here a recurrent neural network structure is leveraged to classify new instances by comparison with existing data in memory \cite{munkhdalai2017meta,santoro2016meta}. Because of similar issues as discussed above, we chose not to go with this technique.
\\
\textbf{Embedding and metric learning approaches} - Here the goal is to project a set of samples into an embedding space such that they become easily distinguishable \cite{koch2015siamese,snell2017prototypical,vinyals2016matching}. This technique is more efficient for occupied/unoccupied instance detection than the two discussed above, in terms of samples and complexity of neural network models. However, To learn efficient projections onto the embedding space, the number of supervised training examples is still significantly higher compared to our expected standards for privacy preserving occupancy detection (need labelled examples in the order of single digits, and not involving transfer learning from other types of data). 
\\
\textbf{Few shot clustering} - This is a relatively new paradigm in application to deep neural networks as major successes have been mainly achieved with supervised classification tasks. UMTRA \cite{khodadadeh2018unsupervised} takes advantage of domain specific augmentations and innate statistical diversity of the data for generating training and validation data. AAL \cite{antoniou2019assume} again takes advantage of data augmentations in unlabeled support sets To generate query data (more details on support set and query data in following section) CACTUs \cite{hsu2018unsupervised} takes advantage of progressive clustering and leverages feature level representations followed by training in an episodic fashion. However we carry out concurrently, progressive clustering and episodic training much like ``E" step and ``M" step in expectation maximization technique as parts of two processes in the clustering that facilitate each other. A similar idea to ours has only been observed very recently in \cite{ji2019unsupervised}, however we make our framework significantly simpler by using siamese neural networks that learn distance functions as a difference in embeddings produced for a pair of images. Also in contrast, We do not apply any explicit statistical distance metrics \cite{irani2016clustering} to our learning algorithm. Additionally, we provide a full schematic on how to apply our technique in a most effective manner for occupancy detection in indoor images.
\\
\textbf{Image based occupancy detection in indoor environment}
\\
Recently deep learning based object detection frameworks have been studied to a great extent including cases for pedestrian detection and person reidentification from several extensively annotated datasets, which has shifted the focus from the much easier task of zonal occupancy detection. However there is still huge scope in developing algorithms that can detect occupancy status by training from minimal number of images while generalizing to different room configurations, poor image qualities and lighting conditions. Existing literature on using deep learning for indoor image based occupancy detection with privacy and sample efficiency concerns is scarce, although a few related applications exist. In \cite{callemein2019anyone}, the authors propose to use omnidirectional image sensors on smart embedded low resolution systems to count occupancy in office spaces. Due to lack of training data like theirs, the authors use available state of the art person detection frameworks to generate annotated samples. We propose to learn zonal occupancy detection on a simple siamese convolutional neural network (CNN) model by self supervised clustering thus avoiding the need to generate labels using existing computationally expensive frameworks. We also use a very low cost and low power camera which is not omnidirectional. In \cite{acharya2018real}, the authors propose deep learning based parking lot occupancy detection. An extensive labelled data set is used for this approach and privacy preservation is not a main concern. In \cite{zou2017occupancy}, the authors propose occupancy detection in office spaces by analyzing large volumes of previously recorded surveillance videos. 
\\
Therefore, based on our survey, it is clear that existing few shot based learning techniques are still not ripe to be deployed efficiently in indoor occupancy detection, and several existing deep learning techniques for image based occupancy detection do not constrain themselves with size of training data, model simplicity for easy deployment on low cost devices and privacy of its occupants~\cite{saha2019occupancy}. The solutions proposed in this paper allow future research to pursue newer ways to tackle these issues. 

\section{Occupancy detection framework}
\begin{figure*}
    \centering
    \includegraphics[width=.85\linewidth]{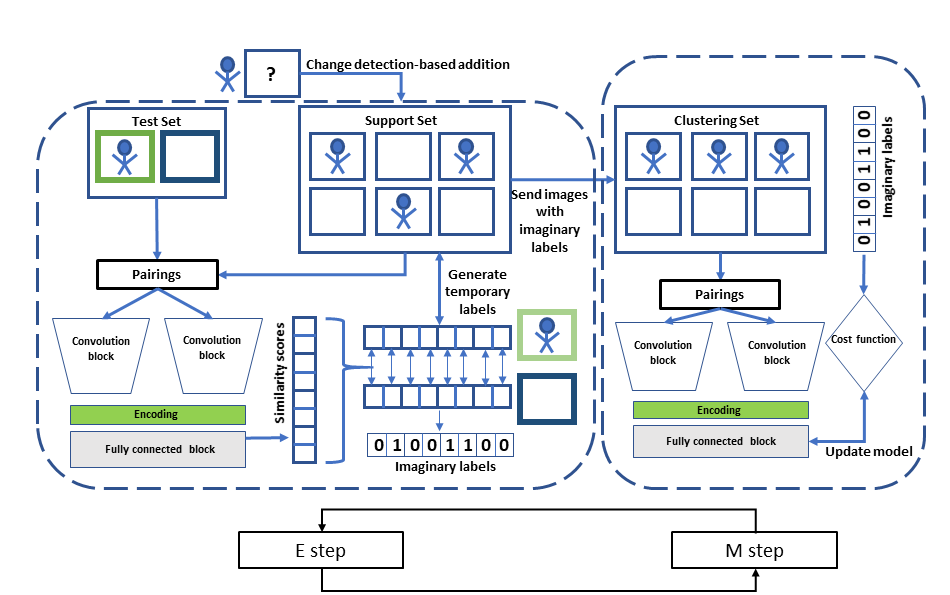}
    \caption{{\bf Schematic of our few shot occupancy clustering algorithm- Siamese Convolutional Few shot Clustering (SCFC). An initial calibration data of a few instances of occupied and unoccupied (test set) is provided to set the target clustering semantic. After that an online adaptation procedure follows where new unlabeled low resolution images are added to database (support set) which is used to compare with initial calibration data and eventually learn to cluster occupied and unoccupied instances without any external supervision. This is achieved by episodic estimation of imaginary labels and readjustment of convolutional network parameters that estimate the difference in a pair of images. For complete algorithm, see \ref{completealgo}}}
    \label{fig:1framework}
\end{figure*}

We propose Siamese Convolution Few shot Clustering \textbf{(SCFC)}. Our few shot episodic clustering based occupancy detection algorithm is summarized in figure~\ref{fig:1framework}. There are several parts to our algorithm and we explain each aspect briefly in the following subsections.
\subsection{Siamese Convolutional Neural Network}
A Siamese convolutional neural network~\cite{varior2016gated} is designed to accept two distinct inputs from each of the two convolutional neural network blocks that map their respective images into a common encoding space. This leads to a function that can compute a metric between the highest-level representations of features on each side. The name ``siamese" comes from the fact that both the CNNs share the same parameters. First this strategy ensures consistency of predictions because it is very unlikely that similar images get mapped to different feature representations, and second, the network is symmetric. In most applications, the networks need to develop a certain bias in expressing input data such that mapped feature representations become easily separable. This is generally done by having the network run through numerous pairs of weakly-supervised data (possibly from scenarios other than those in deployment) that are labelled as either ``similar" or ``dissimilar". After proper weights to the encoding space have been learned a simple fully connected neural network or even Support Vector Machine (SVM) or linear regression based methods can be trained to classify the joint encoding as members of different unique classes in the data collected during deployment. In our experiments we use a two layer convolution block as the weights of the Siamese network, followed by a three layer fully connected neural network block as the classifier. The Siamese classifier takes as input two images and predicts a probability $p \in [0,1]$. Values close to 0 implicitly means that images are similar and close to 1 means they are highly dissimilar. This value is used as a metric for subsequent episodic clustering (described below).

\subsection{Training setup and some nomenclature}
The training data is presented to our few shot classifier arranged in three different folders:
\begin{enumerate}
    \item \textbf{Test Set (clustering target or calibration set)}: This is the few handpicked occupied and unoccupied images labelled as 1 (occupied) and 0 (unoccupied). Number of handpicked examples is limited to at most 5 occupied and 5 unoccupied images for a certain zone. These few labelled images are used to decide the clustering semantic. In our case, it is occupancy detection, however it can be easily adjusted to perform any standard few shot classification task (explained later in results section).
    \item \textbf{Support Set (unlabeled)}: This consists of all unlabeled data collected continually from a certain zone. Images could represent either ``occupied" or ``unoccupied" instances. There is no restriction on the number of images required and the class distribution. We assume that under general operational circumstances a simple change detection algorithm can be used to trigger image collection from the zone by the camera, thus updating the images stored in ``support set" in real time.
    \item \textbf{Clustering Set}: This is the part of the data that is used to train our Siamese encoder to classify a pair of images as ``similar" or ``dissimilar". Images in this folder are selected from the support set in an episodic fashion based on certain clustering rules (described in ``E" step) and assigned imaginary labels (imaginary because they are decided based on the encoding of the Siamese CNN) of 1 (occupied) and 0 (unoccupied). Images in this folder can again have any distribution of classes and any number of examples. Note that the only cue provided to the CNN about the nature of distinction required in the clustering is through the extremely small number of true labels in Test Set, because at the start of the training clustering set is initialized to be the same as the test set. If we would have used a normal unsupervised clustering algorithm on all images in support set, there are chances that the images may be even clustered based on undesirable conditions such as lighting conditions. Thus combining few shot learning with episodic clustering we are trying to achieve ``targeted-clustering".
\end{enumerate}
We randomly initialize the weights of the neural network, and pretrain it for $n$ steps on all pairs generated from the Test Set. After that we start our episodic few shot learning and clustering algorithm for $N$ steps. Here $n << N$. The goal of the pretraining is to obtain good starting weights for the Siamese encoder. In the next subsections we describe our algorithm in more detail.

\subsection{The ``E'' step}\label{Estep}
Let $s \in S$ be each image in the current support set $S$. Let $SPT$ denote all pairs of images where one of them is taken from support set and one of them is taken from test set, and $spt \in SPT$. Let $p_{SPT}$ be the list of all probabilities obtained by operating the Siamese classifier on each pair $spt \in SPT$. Now $p_{SPT}$ can be arranged in the form of two (wide) tables:

\textbf{Occupied table}: Here each row represents the probabilities with one element of the pair being in the occupied images in the test set, and each column represents the probabilities with one element of the pair being an image in the support set. 

\textbf{Unoccupied table}: Here each row represents the probabilities with one element of the pair being in the unoccupied images in the test set, and each column represents the probabilities with one element of the pair being an image in the support set. 

Now after arranging the probabilities in a tabular form, we perform mean thresholding for each row. If the probability value is greater than the row mean the element in the table is set to value 1 and if the probability value in the table is less than or equal to the mean of the probabilities in that row, that element in the table is set to a value of 0. Finally, we chose each column in both the tables. If the average score of the column is greater in occupied table than in unoccupied table, then we assign an imaginary label of 1 (occupied) to that corresponding image in the support set and save a copy of that image in the clustering set along with the imaginary label. If the average score of the column is greater in unoccupied table than in occupied table, then we assign an imaginary label of 0 (unoccupied) to that corresponding image in the support set and save a copy of that image in the clustering set along with the imaginary label. After the ``E'' step we switch over execution to the ``M'' step.
\subsection{The ``M'' step}\label{Mstep}
In this step we train the entire Siamese CNN based one shot classifier on the imaginary labelled data in the clustering set. 
\\
\textbf{Loss function}: Let the minibatch size be represented by $M$, and $i$ indexes the $i^{th}$ minibatch. Now $y(x_1^i,x_2^i)$ is a length-$M$ vector containing the labels in the minibatch, where we already mentioned earlier that $y(x_1^i,x_2^i) = 1$ whenever $x_1$ and $x_2$ are from the same class (that is, both are occupied or both are unoccupied instances), and $y(x_1^i,x_2^i) = 0$ otherwise. Also, here $p$ is the output of the sigmoid activation of the last layer of the classifier as mentioned earlier too. We use a regularized cross-entropy objective function formulated just as in case of a binary classifier as follows:
\begin{equation}
\begin{split}
    L(x_1^i,x_2^i) = y(x_1^i,x_2^i)log p(x_1^i,x_2^i) + &
    \\ 
    (1-y(x_1^i,x_2^i))log(1-p(x_1^i,x_2^i)) + \Lambda^T(|w|)^2
\end{split}
\end{equation}
Here $\Lambda$ are the regularization terms and $w$ is the weights matrix. Based on the above loss function we perform a single step of minibatch gradient descent using all of the samples in the support set along with their corresponding imaginary labels. After this update we switch back to the ``E'' step. This alternating ``E'' step and ``M'' step is carried out in each episode until convergence.

\subsection{Complete Algorithm}\label{Completealgo}
The complete few shot occupancy clustering algorithm is provided in algorithm~\ref{completealgo}. Here \textbf{RANDOM-PAIR} is a function that extracts a random pair of images from a set of images with replacement, \textbf{LABEL} is a function that returns either the assigned true label or the assigned imaginary label of an image, \textbf{ROW-MEAN}, and \textbf{COLUMN-MEAN} provides the row mean and column means of input table indexed with the rows or column numbers.

\begin{algorithm}
\caption{Few shot occupancy clustering}
\label{completealgo}
\begin{algorithmic}[1]
  \REQUIRE Test Set ($\mathcal{T}=\{\mathcal{T}_1,\dots \mathcal{T}_k$\}), Support Set ($\mathcal{S}=\{\mathcal{S}_1,\dots \mathcal{S}_K$\}), and Clustering Set ($\mathcal{C}=\{\}$). $\mathcal{T}=\mathcal{T}_O \bigcup \mathcal{T}_U$. 
  \REQUIRE All images in $\mathcal{T}_O$ are of occupied instances and all images in $\mathcal{T}_U$ are of unoccupied instances.
    \STATE Initialize Siamese Convolutional Neural Network ($F$) with random weights $\theta$
    \newline
    \COMMENT {Pretrain (warmstart) $F$ on $\mathcal{C}$ for $n$ epochs}
    \FOR{$i=0;i<n;i\gets i+1$}
        \STATE $x_1^i, x_2^i \gets$ RANDOM-PAIR$(\mathcal{T})$
        \STATE $p \gets F(x_1^i,x_2^i,\theta)$
        \STATE $y\gets 1$ if LABEL($x_1^i$) = LABEL($x_2^i$) else 0
        \STATE $L(x_1^i,x_2^i,\theta) = ylog p +(1-y)log(1-p)+\Lambda^T(|\theta|)^2$
        \STATE $\theta \gets$ SGD$(L(x_1^i,x_2^i,\theta))$
    \ENDFOR
    \newline
    \COMMENT {Episodic clustering begins, $1<N<\infty$}
    \STATE Initialize counter c = 0
    \newline
    \COMMENT{E-step(11-19), M-step(20-27)}
  \WHILE{$c < N$} 
      \STATE $P_O\gets \{F(x_1^i,x_2^i,\theta) \forall (x_1^i \in \mathcal{T}_O \bigwedge  x_2^i \in \mathcal{S})\} :
      P_O(i,j) = \{F(x_1^i,x_2^j,\theta) : (x_1^i = \mathcal{T}_{O_i}\bigwedge  x_2^j = \mathcal{S}_j)\}$
      
      \STATE $P_U\gets \{F(x_1^i,x_2^i,\theta) \forall (x_1^i \in \mathcal{T}_U \bigwedge  x_2^i \in \mathcal{S})\} :
      P_U(i,j) = \{F(x_1^i,x_2^j,\theta) : (x_1^i = \mathcal{T}_{U_i} \bigwedge  x_2^j = \mathcal{S}_j)\}$
      
      \STATE $P_O(i,j) \gets 1$ if $P_O(i,j)$ $>$ ROW-MEAN($P_O(i,)$) else 0
      
      \STATE $P_U(i,j) \gets 1$ if $P_U(i,j)$ $>$ ROW-MEAN($P_U(i,)$) else 0
      
      \IF{COLUMN-MEAN($P_O(,j)$) $>$ COLUMN-MEAN($P_U(,j)$)}
        \STATE LABEL($S_j$)$\gets 1$
      \ELSE
        \STATE LABEL($S_j$)$\gets 0$
      \ENDIF
      \STATE $\mathcal{C}\gets \mathcal{C}\bigcup$ $\{ (S_j,$ LABEL($S_j$) $)\}$
     \FOR{$X_i \in \{(\mathcal{S}_i,\mathcal{S}_j) \forall \mathcal{S}_i \in \mathcal{S} \bigwedge \mathcal{S}_j \in \mathcal{S} \}$  }
        \STATE $x_1^i, x_2^i \gets X_i$
        \STATE $p \gets F(x_1^i,x_2^i,\theta)$
        \STATE $y\gets 1$ if LABEL($x_1^i$) = LABEL($x_2^i$) else 0
        \STATE $L(x_1^i,x_2^i,\theta) = ylogp +(1-y)log(1-p)+\Lambda^T(|\theta|)^2$
        \STATE $\theta \gets$ SGD$(L(x_1^i,x_2^i,\theta))$
    \ENDFOR
       \COMMENT{ Use Structural Similarity Index metric (SSIM)~\cite{brunet2011mathematical} to detect any changes in image}
      \STATE $\mathcal{S} \gets \mathcal{S}\bigcup$ SSIM-CHANGE()
      \STATE $c \gets c+1$
  \ENDWHILE
\end{algorithmic}
\end{algorithm}

\section{Evaluation on Benchmarks}\label{benchmarks}

\subsection{Dataset on occupancy detection}
The goal of our algorithm is room occupancy estimation with almost unlabeled data by online adaptation to changing scenes using few shot clustering. We found a lack of established benchmarks for few shot occupancy clustering as most approaches focus on bounding box style object detection applied to person detection for occupancy estimation. We therefore decided to apply our algorithm on the open sourced ``People in Indoor ROoms with Perspective and Omnidirectional cameras" (PIROPO) dataset. This dataset consist of video frames collected in an indoor shared open area and in a computer lab, under different lighting conditions, different number of occupants and changing scene background. These different settings provide the perfect testbed to demonstrate the robustness of our algorithm under the limitation of number of training data. We randomly select at most 5 instances of occupied and unoccupied images for each room to start our clustering on PIROPO. Results obtained in table \ref{benchmark-table1} are compared with the k nearest neighbor clustering (kNN) algorithm which is also a similarity metric based clustering technique. Most of the proposed state of the art approaches~\cite{vinyals2016matching,snell2017prototypical,ravichandran2019few} typically train their networks to develop an inductive bias by constructing few shot learning tasks from several images from different training classes and then transferring the abilities to perform few shot classification on a new test class. In the spirit of learning from level 0, without a pretrained inductive bias and leveraging the abilities of clustering to form self supervised examples, we believe kNN, although a purely unsupervised shallow model is one of the few techniques that can serve as a fair comparison to our technique.

\subsection{Adaptation to general pairwise classification}
The second dataset that we chose is the MNIST handwritten digit dataset. Though MNIST is not an occupancy dataset, our main intention here is to show that our algorithm is not only limited to occupancy detection, but also on other clustering problem as well. To model the MNIST dataset for our purpose, we segregate the dataset into multiple pairs of digits, for example, pairing class 0 and 1, class 2 and 3, etc. Like PIROPO we compare the results with kNN clustering in table \ref{benchmark-table1}.

\begin{table}
\centering
\begin{tabular}{c|c|c|}
\cline{2-3}
                                    & \multicolumn{2}{c|}{\textbf{Accuracies (\%)}} \\ \hline
\multicolumn{1}{|c|}{\textbf{Dataset}}       & \textbf{Our Algo.}        & \textbf{kNN}          \\ \hline

\multicolumn{1}{|c|}{PIROPO}        & 82.5                 & 78.50         \\ \hline
\multicolumn{1}{|c|}{MNIST 0 and 1} & 86.55            & 84.24              \\ \hline
\multicolumn{1}{|c|}{MNIST 2 and 3} & 82.64            & 79.79              \\ \hline
\multicolumn{1}{|c|}{MNIST 4 and 5} & 83.25            & 82.61     \\ \hline
\multicolumn{1}{|c|}{MNIST 6 and 7} & 84.73            & 80.15              \\ \hline
\multicolumn{1}{|c|}{MNIST 8 and 9} & 81.02              & 79.00        \\ \hline
\end{tabular}
\vspace{0.2cm}
\caption{Comparison of performance of our algorithm with k-nearest neighbor clustering in PIROPO dataset and MNIST pairwise digit clustering}
\label{benchmark-table1}
\end{table}

\begin{table}
\centering
\begin{tabular}{|l|l|}
\hline
\textbf{Model} & \textbf{5-way 5-shot Acc.} \\ \hline
UMTRA $~\cite{khodadadeh2019unsupervised}$ & 50.73\% \\ \hline
CACTUs-MAML $~\cite{hsu2018unsupervised}$ & 53.97\% \\ \hline
Matching Nets  $~\cite{vinyals2016matching}$ & 55.3\% \\ \hline
MAML $~\cite{finn2017model}$ & 63.1\% \\ \hline
Meta-learn LSTM $~\cite{ravi2016optimization}$ & 63.1\% \\ \hline
Centroid networks $~\cite{huang2019centroid}$ & 64.5\% \\ \hline
\textbf{SCFC (ours)} & 65.5\% \\ \hline
\end{tabular}
\caption{MiniImageNet few-shot classification accuracies (95\% confidence intervals)}
\label{benchmark}
\end{table}

\subsection{Adaptation to few shot classification benchmarks}
Few shot occupancy detection from highly obfuscated images open the door to more advanced indoor image recognition tasks such as activity recognition and privacy preserving patient health monitoring. To motivate that our framework can be adapted to different kinds of tasks we demonstrate performance on the benchmark miniImageNet dataset on 5-way 5-shot classification. Since our algorithm focused earlier on detecting whether spaces are occupied or not (2-way 5-shot classification) we have to make minor modifications. We chose 5 random classes from the miniImageNet dataset and trained a one vs. all classifier for each of the class using our few shot clustering technique. Thereafter we combined each of the learned clusters using simple majority voting technique to obtain separate clusters for each of the 5 different classes. Table~\ref{benchmark} aggregates the previously reported performance on past several techniques (supervised,unsupervised, meta-learning) with the bottom row reporting our performance on the 5-way 5-shot task. Our technique \textbf{(SCFC)} achieves comparable accuracy to recently proposed centroid networks~\cite{huang2019centroid} which is also based on unsupervised few shot clustering. 

\begin{figure*}[!thb]
    \centering
    \subfigure[Occupied (Room 1)]{\label{fig: occupied_1}\includegraphics[width=.225\linewidth]{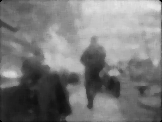}}
    \subfigure[Unoccupied (Room 1)]{\label{fig: unoccupied_1}\includegraphics[width=.225\linewidth]{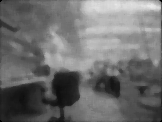}}
    \hspace{0.5cm}
    \subfigure[Occupied (Room 2)]{\label{fig: occupied_2}\includegraphics[width=.225\linewidth]{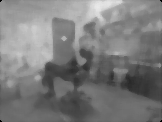}}
    \subfigure[Unoccupied (Room 2)]{\label{fig: unoccupied_2}\includegraphics[width=.225\linewidth]{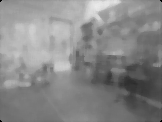}}    
    
    \subfigure[Accuracy (Room 1)]{\label{fig: accuracy_1}\includegraphics[width=.45\linewidth]{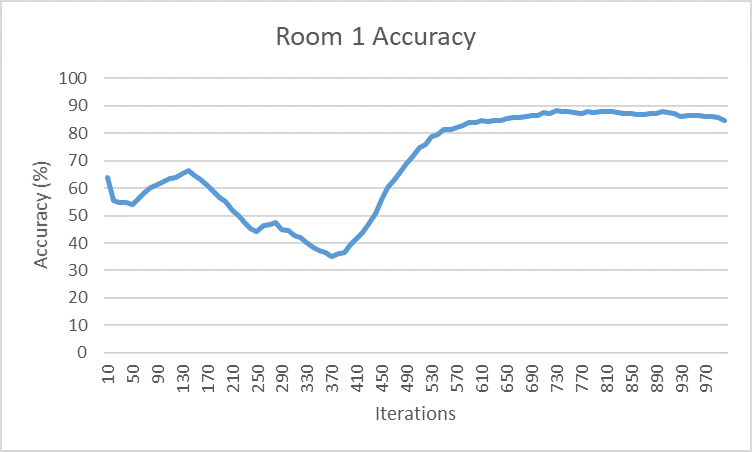}}
    \hspace{0.5cm}
    \subfigure[Accuracy (Room 2)]{\label{fig: accuracy_2}\includegraphics[width=.45\linewidth]{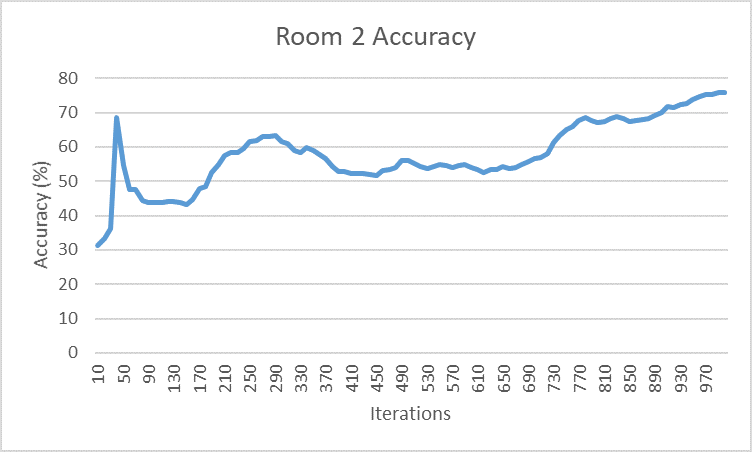}}    
    
    \subfigure[Occupied (Room 3)]{\label{fig: occupied_3}\includegraphics[width=.225\linewidth]{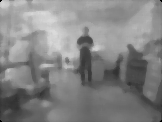}}
    \subfigure[Unoccupied (Room 3)]{\label{fig: unoccupied_3}\includegraphics[width=.225\linewidth]{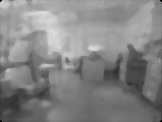}}
    \hspace{0.5cm}
    \subfigure[Occupied (Room 4)]{\label{fig: occupied_4}\includegraphics[width=.225\linewidth]{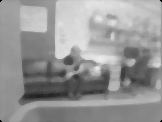}}
    \subfigure[Unoccupied (Room 4)]{\label{fig: unoccupied_4}\includegraphics[width=.225\linewidth]{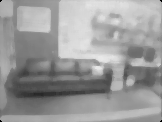}}    
    
    \subfigure[Accuracy (Room 3)]{\label{fig: accuracy_3}\includegraphics[width=.45\linewidth]{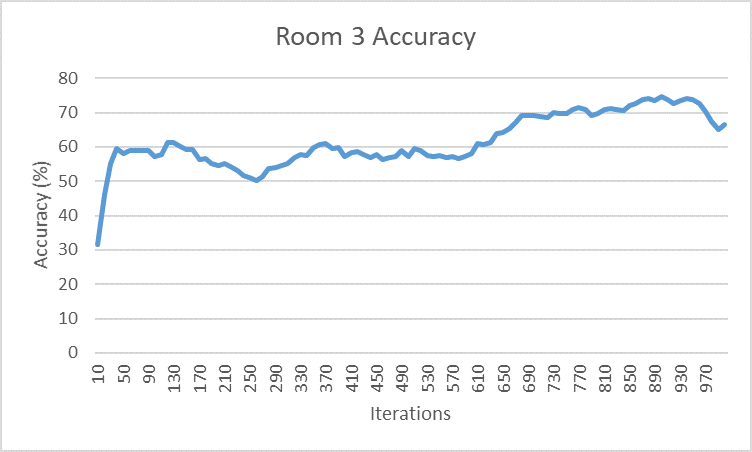}}
    \hspace{0.5cm}
    \subfigure[Accuracy (Room 4)]{\label{fig: accuracy_4}\includegraphics[width=.45\linewidth]{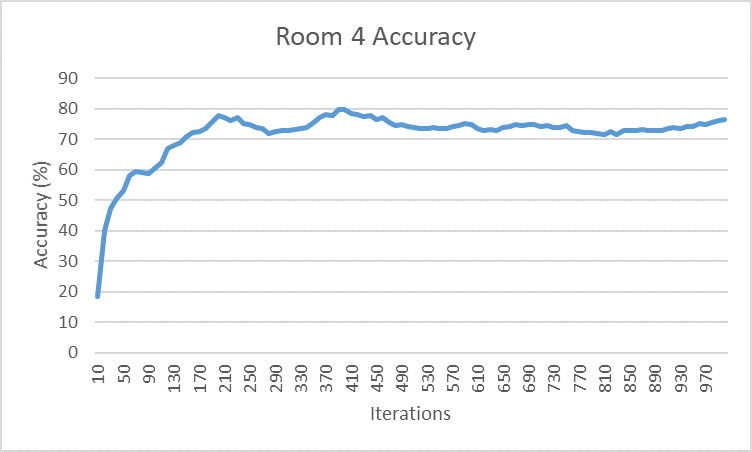}}    
    \caption{{\bf Figure showing provided starting single pair annotations (occupied and unoccupied) for each room in our collected dataset and corresponding accuracies obtained with our few shot clustering algorithm over self supervised clustering epochs. Note that the accuracies do not start from 0 as we use the few available ground truth pairs to ``warm start" the clustering algorithm by training the Siamese neural networks on them for a very few number of epochs.}}
    \label{fig: accuracy with example image}
\end{figure*}

\section{Hardware, data collection and experiment results}
\subsection{Low power camera images}
We have developed battery-free cameras able to collect low-quality privacy preserving images. The battery-free video streaming camera in \cite{camera} uses backscatter communication technology \cite{abc} to send the recorded images to a receiver. 
Backscatter is an low power communication technology wherein a transmitter generates an RF (Radio Frequency) excitation signal, the sensor modulates the excitation signal reflection of its antenna, and the receiver listens to the reflection to decode the sensor data. Since the sensor does not need to generate any signal to communicate with the receiver and it only reflects or absorbs the RF signal generated by the transmitter the sensor can operate with harvested ambient light and RF energy. The sensor (camera) has a dual harvester board capable of harvesting the RF energy generated by the transmitter and also harvesting energy from ambient light using small solar cells. 

\begin{figure}[!thb]
    \centering
    \subfigure[Image sensor]{\label{fig: hardware 1 a}\includegraphics[angle= -0.8, width=.27\linewidth]{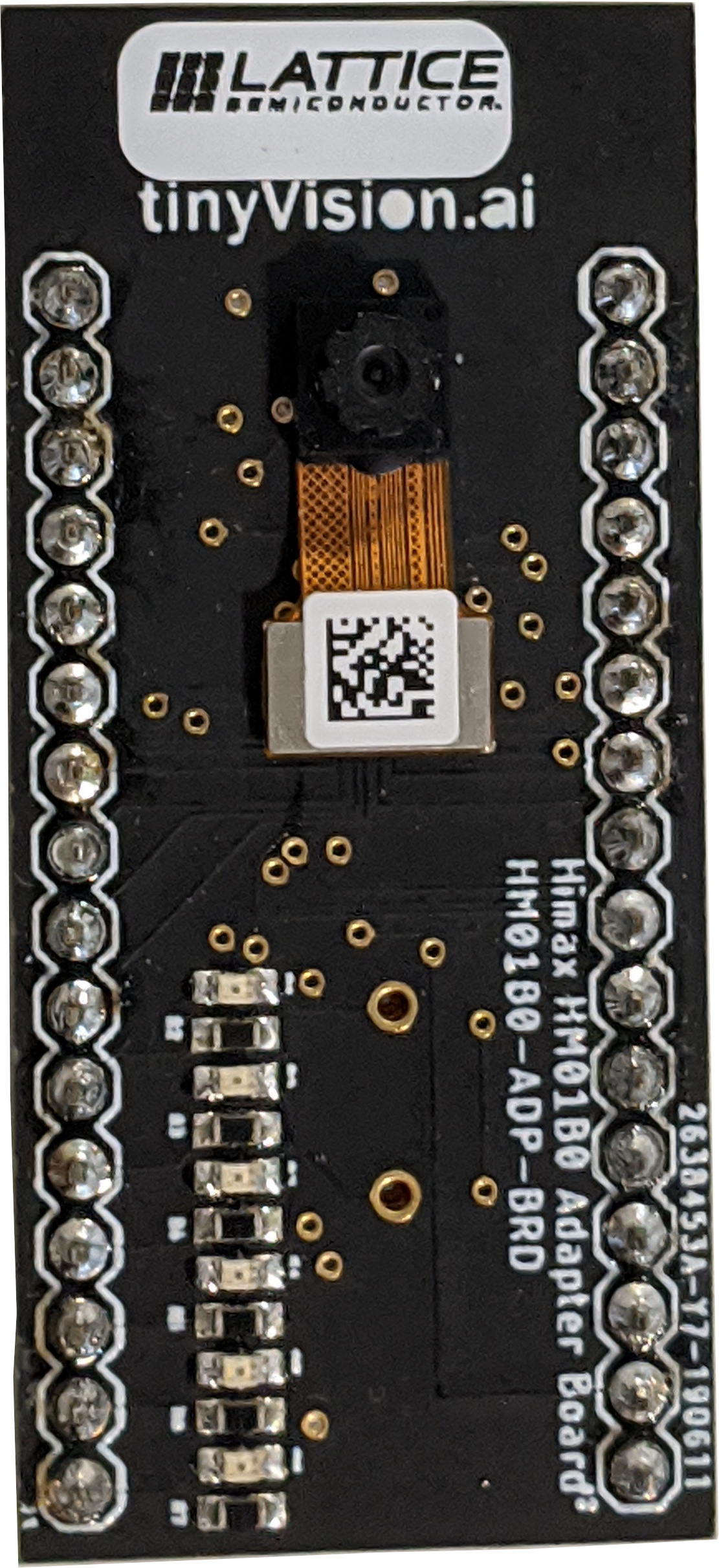}}
    \hspace{1cm}
    \subfigure[Controller]{\label{fig: hardware 1 b}\includegraphics[width=.37\linewidth]{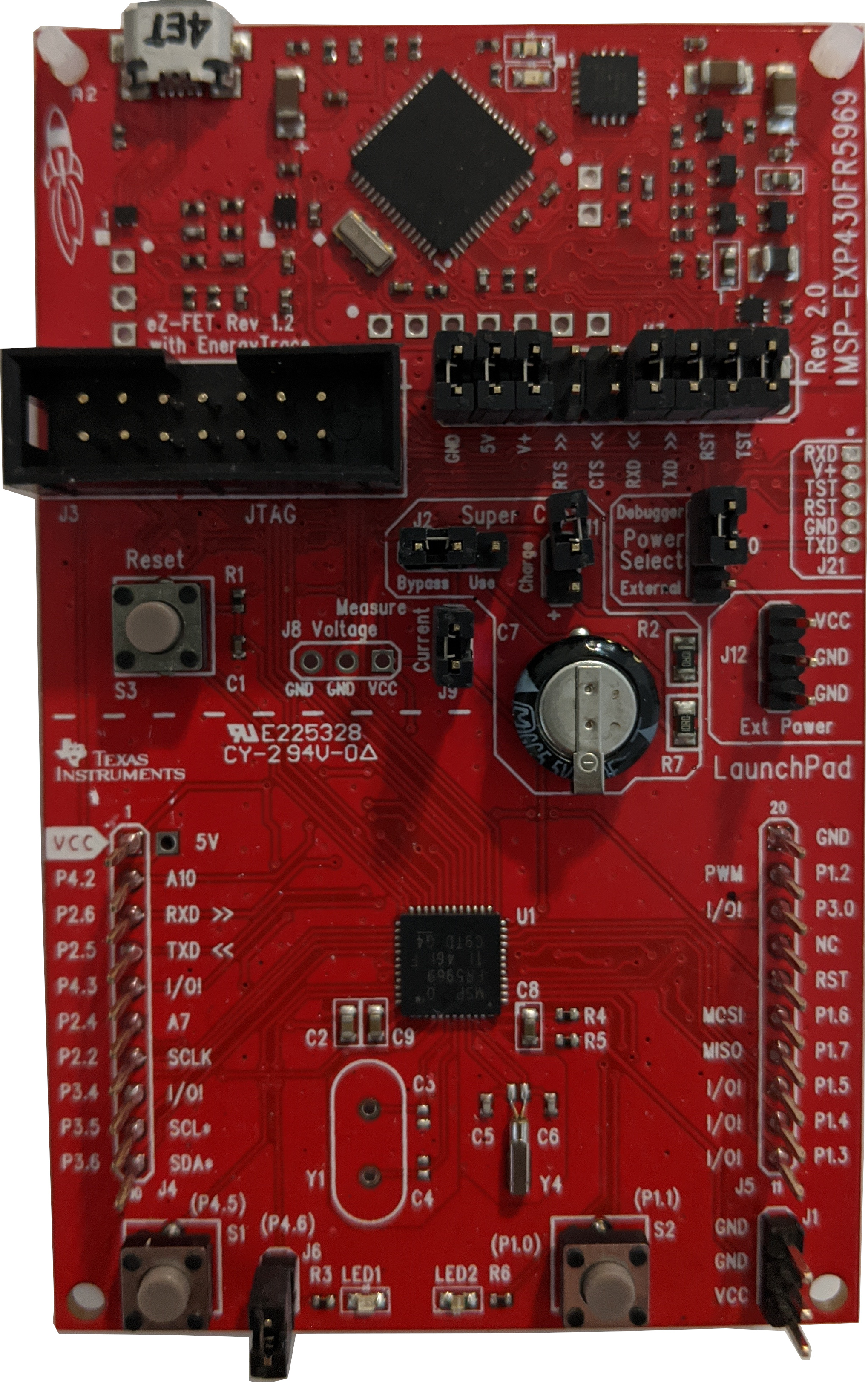}}
    \caption{{\bf Development boards of the image sensor and controller.}}
    \label{fig: hardware 1}
\end{figure}

We use a Himax HM01B0 image sensor, shown in Fig~\ref{fig: hardware 1 a} to collect our 160*120 pictures. This ultra low power image sensor is capable of recording QVGA (320*240) and QQVGA (160*120) images; the power consumption for these operation modes is 2 mW, and 1.1 mW, respectively. A Texas Instruments MSP430FR5969 micro-controller, shown in Fig~\ref{fig: hardware 1 b},  initializes the image sensor and receives the recorded digital pixel values. 

Our system can transfer the images using a wireless backscatter communication system. The operation is battery-free and powered by the RF and ambient light harvested energy. We make our collected data publicly accessible. We have performed data collection on 6 different residential settings under various lighting conditions and number of people occupying the place. We denote the data from each room as $Room_i$ where $i$ is the assigned room number where the data was collected. After collecting the images, median filters are applied on each image to partially recover any dropped pixel values due to packet loss during image transfers. Some resulting images are shown in Figure~\ref{fig: accuracy with example image}(a-d) and (g-j). The images shown appear to be a little blurry after median filtering, but in the same time, it also serves the purpose of privacy preserving, which is one of the main concern of using an image modality in occupancy detection.

\subsection{Privacy preservation factor- maxpooling window size}
\begin{figure}
  \centering
  \includegraphics[width=0.50\textwidth]{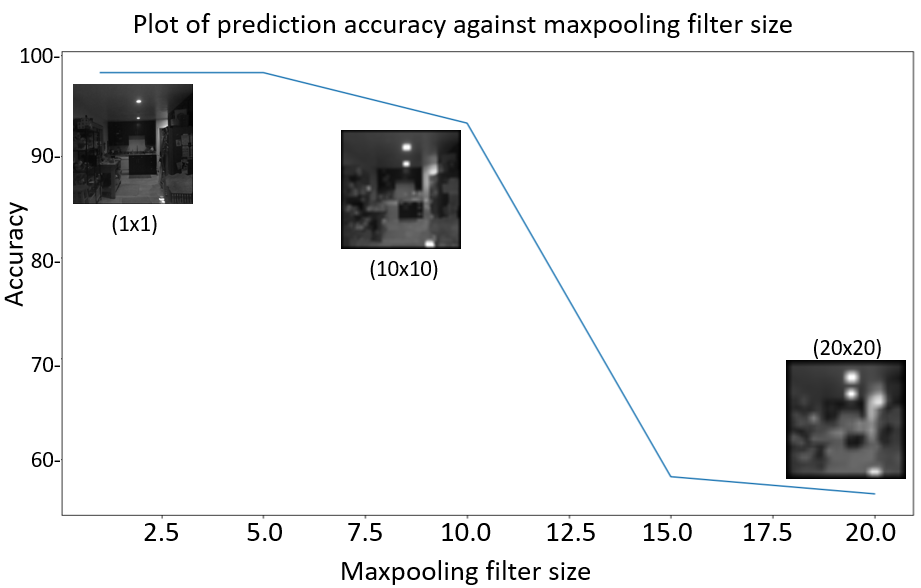}
  \caption{Image of an unoccupied kitchen at various resolutions showing tradeoff between selected pooling window size and test occupancy measure accuracy with a cutoff of 0.5. Increasing pooling window size leads to more privacy, but destroys prediction capability beyond a size of 20x20}
  \label{fig: acc_vs_filter_size}
\end{figure}

Based on the energy available to the device and distance between receiver and transmitter, image quality collected from our hardware can be affected. We collected images from normal camera and tried to simulate this varying amount of degradation by using maxpooling filters. like the maxpooling operation in convolutional neural networks, here we degrade the resolution of an image by selecting only the maximum element in every small patch of the image (obtained by moving the maxpooling window over the image with a stride length of 1 pixel) and duplicating that value over all the elements selected over that window. For an initial image size of 112x112 pixels we increase the maxpooling window size from 1x1 pixel (no maxpooling) to 20x20 pixel (severe degradation). We observe that significant privacy can be maintained while maintaining considerable accuracy close to a maxpooling window size of 10x10 pixels (see figure~\ref{fig: acc_vs_filter_size})



\subsection{Experiment results}
In this section, we will present our algorithm's results on the dataset collected by our hardware. For algorithmic baseline comparison, we have selected the k-Nearest Neighbor(kNN) algorithm, which is also a similarity-metric based algorithm. 
For the training data, we selected a maximum of 5 images from each occupied and unoccupied state to feed into our algorithm. The images selected in the test set and the resulting clustering accuracies along with the proceeding of number of epochs in shown in Figure~\ref{fig: accuracy with example image}(a-d) and (g-j). The impact of limited training data is shown clearly in almost all the dataset in Table \ref{experiment-results}, where kNN is only able to achieve accuracies $\sim$65\% to $\sim$70\% ( except for MNIST dataset), while our algorithm improves the accuracies by $\sim$10\%. 
For the various datasets we perform previously described alternating ``E-M'' style clustering and learning for up to 1000 epochs. A more detailed view of the training process are shown in Figure~\ref{fig: accuracy with example image}, where it illustrates the increasing trend in performance over the training iterations.

\begin{table}
\centering
\begin{tabular}{c|c|c|}
\cline{2-3}
                                    & \multicolumn{2}{c|}{\textbf{Accuracies (\%)}} \\ \hline
\multicolumn{1}{|c|}{\textbf{Dataset}}       & \textbf{Our Algo.}        & \textbf{kNN}          \\ \hline
\multicolumn{1}{|c|}{Room 1}         & 84.79            & 65.50         \\ \hline
\multicolumn{1}{|c|}{Room 2}         & 76.29            & 69.00         \\ \hline
\multicolumn{1}{|c|}{Room 3}         & 74.60            & 63.50         \\ \hline
\multicolumn{1}{|c|}{Room 4}         & 76.61            & 69.50         \\ \hline
\multicolumn{1}{|c|}{Room 5}         & 76.62            & 70.32         \\ \hline
\multicolumn{1}{|c|}{Room 6}         & 85.84            & 71.47         \\ \hline

\end{tabular}
\vspace{0.2cm}
\caption{Comparison of performance of our algorithm with k-nearest neighbor clustering over all 6 different room occupancy dataset collected by us.}
\label{experiment-results}
\end{table}

\section{Conclusion}
We presented a novel algorithm for few shot learning without transfer of pretrained knowledge by leveraging a clustering technique. We motivated an example use case of our algorithm for few shot privacy preserving occupancy detection in residential buildings from images collected by novel low power hardware introduced along with the algorithm. We collect occupied/unoccupied images under several different room and lighting configurations from our novel hardware and release the dataset. On analyzing performance of our algorithm on the data collected by our hardware it is seen that reasonable accuracies are achieved for occupancy detection starting with labelled images in the order of single digits. We also demonstrated generalization of our algorithm to other different datasets and scenarios, such as on the PIROPO dataset, pairwise clustering of MNIST digits, and the few shot classification benchmark miniImageNet provided single digit actual labels. In the future we hope to achieve improved performance on novel clustering tasks which have relevant applications in occupancy detection and hope to surpass existing supervised learning benchmarks with our few shot learning scheme employing only few single digit true labels.

{\small
\bibliographystyle{ieee_fullname}
\bibliography{egbib}
}

\end{document}